# On Constraint-Based Lambek Calculi

Jochen Dörre

*Institut für Machinelle Sprachverarbeitung (IMS)*
*University of Stuttgart, Germany*
*email:* Jochen.Doerre@ims.uni-stuttgart.de

and

Suresh Manandhar

*Language Technology Group*
*Human Communication Research Centre*
*University of Edinburgh, Scotland*
*email:* Suresh.Manandhar@ed.ac.uk

**Abstract.** We explore the consequences of layering a Lambek proof system over an arbitrary (constraint) logic. A simple model-theoretic semantics for our hybrid language is provided for which a particularly simple combination of Lambek's and the proof system of the base logic is complete. Furthermore the proof system for the underlying base logic can be assumed to be a black box. The essential reasoning needed to be performed by the black box is that of *entailment checking*. Assuming feature logic as the base logic entailment checking amounts to a *subsumption* test which is a well-known quasi-linear time decidable problem.

## 1 Background

In recent years there has been a growing awareness of a need to design grammar logics that incorporate both the resource-mindedness of categorial grammars and the typed constraint-based approach of HPSG [15] [16]. We believe that the long-term goal of this enterprise is to provide an incremental and largely deterministic model of sentence comprehension within a constraint-based setting — something that current HPSG lacks. The constraint-based basis is important, since this provides an excellent knowledge representation, engineering and structuring environment for NLP.

Although, there has been a lot of work on finer systems of categorial grammars [11] [12] none directly build on a unification based framework. In a system such as that of Moortgat's [10] categories can be built using tuples containing *type*, *syntax* and *structure* in the spirit of UCG [23]. However, feature descriptions cannot be employed for instance to describe the syntax. Thus the consequences of a more direct and perhaps more pragmatic approach remains unexplored. In this paper, this is precisely what we do.

UCG [23] and CUG [21] provided the initial integration of such an approach. However, both lacked a rigorous model-theory. Even worse, their underlying unification-based proof system is incompatible with the straightforward model theory of feature-based categorial types (where feature terms take the place of basic categories like in $(cat{:}s)/(cat{:}np \& case{:}acc)$), when making the following two basic assumptions, which we believe are uncontroversially desirable in such a theory.

1. The denotation of complex types should be composed in the same way as in standard semantics for (Lambek) categorial systems. No matter whether we use string semantics or ternary frame semantics [6], types denote sets of objects and a complex type $A/B$ or $B\backslash A$ denotes a left-, resp. right-residual w.r.t. a join operation (or relation) $\odot$, *i.e.*, those 'functional' objects which when joined with a $B$ object (to their left, resp. right) yield an $A$ object.

2. Feature terms should induce a subtype ordering on categories in the sense that more specific types should denote subsets of less specific ones (e.g. $[\![cat\!:\!np\,\&\,case\!:\!acc]\!] \subset [\![cat\!:\!np]\!]$).

Now, this implies that, for instance, $A/B \odot B' \Rightarrow A$ ("when something is the join of $A/B$ and $B'$, then it is of type $A$") if and only if $B'$ is a *subtype* of $B$. On the other hand this sequent is derived in the unification-based systems whenever $B$ and $B'$ unify. Note that there is an essential difference between the two views in that our model theory postulates an asymmetry between $B$ and $B'$ *i.e.* $B'$ is informationally more specific than $B$. In order to refer to one of the two approaches, we will talk of *subsumption-based* vs. *unification-based argument binding*.

Here, we explore the consequences of adding a Lambek proof system over an arbitrary (constraint) logic while following the subsumption-based argument binding approach, *i.e.*, the one evolving naturally from the model-theoretic view. We provide a simple set-theoretic semantics for our hybrid language and show that the Lambek proof system carries over to this hybrid logic and furthermore the proof system for the underlying base logic can be assumed to be a black box. We study primarily the case where the essential reasoning needed to be performed by the black box is that of *entailment checking*. Assuming feature logic as the base logic entailment checking amounts to a *subsumption* test which is a well-known quasi-linear time decidable problem [2] [20].

Section 2 reviews linguistic motivation in favor of the subsumption-based approach, which we borrow from recent work by Bayer and Johnson [3]. The formal framework for our combined logics will be the (easy) extension of Lambek's calculus with subtyping and is presented in Section 3. Next, we define in Section 4 the layered Lambek system over arbitrary base logics (for the description of basic types) as well as the special case, where feature logic is emplyed. Section 5 closes with a discussion on conceivable limitations and the possibility of having both the subsumption-based and the unification-based approaches in one system.

## 2   Double Coordinations

Recently, Bayer and Johnson [3] have given an analysis of agreement phenomena in coordinations, which strongly supports the view of subsumption-based argument binding and which we review here as the primary linguistic motivation for our approach. Consider examples 1a–d.

(1)  a. Kim [$_{VP}$ [$_V$ became ] [$_?$ [$_{AP}$ wealthy ] and [$_{NP}$ a Republican ]]]]

    b. *Kim [$_{VP}$ [$_V$ grew ] [$_?$ [$_{AP}$ wealthy ] and [$_{NP}$ a Republican ]]]]

    c./d. Kim grew [$_{AP}$ wealthy ] / * [$_{NP}$ a Republican ]

Clearly, the contrast between (1a) and (1b) should be explained on the basis that *become* admits AP and NP complements, whereas no NP complement is allowed to follow *grew*. Assuming that we do not want to give coordinations a metagrammatical treatment, but attempt a phrase-structure analysis, then the question arises, what category should be given a coordination like in (1a). In most 'unification-based' approaches to grammar this kind of polymorphic coordination is accounted for by requiring that in a coordination the feature structure of the coodination *subsumes* (or as a stronger condition: is the *generalisation* of) all the feature structures of the coordinated elements. This is captured by a coordination rule like

$$X_0 \longrightarrow X_1 \; conj \; X_2$$
$$\text{where } X_0 \sqsubseteq X_1 \text{ and } X_0 \sqsubseteq X_2$$

(cf. [18]). Due to the additional assumption that types AP, NP, VP, PP are represented with the help of two binary features $\pm v$ and $\pm n$ as $\{+v,+n\}$, $\{-v,+n\}$, $\{+v,-n\}$, $\{-v,-n\}$, respectively, the polymorphic subcategorisation requirement of *become* can be encoded as $+n$. We now may assume category $+n$ also for the coordination in (1a), because this subsumes both AP ($\{+v,+n\}$) and NP ($\{-v,+n\}$). (1b) is correctly ruled out, since *grew* requires the coordination to be $\{+n,+v\}$, and thus it cannot satisfy the subsumption constraint for the NP.

This approach to coordination, however, does not carry over to 'double coordinations' considered in [3], where in addition the verbal part consists of a coordination:

(2)  a. * Kim [ grew and remained ] [ wealthy and a Republican ].

Assuming a standard encoding of subcategorisation information on the verb the coordination rule here would predict a weakening (generalisation) of the subcategorisation requirements of two conjoined verbs (see Fig. 1). With the above rule a phrase "$v_1$ and $v_2$" would admit any complement whose type subsumes the type that $v_1$ selects for as well as the one $v_2$ selects for. This fits with the data as long as the complement(s) have maximally specific types (*i.e.* AP, NP, VP, or PP). However, the analysis breaks down, if the complement is itself a coordination. For such a double coordination construction an analysis using the above coordination rule would lead to a complete relaxation of any subcategorisation requirements (any combination of verbs can take any combination of phrases as complement), since we simply may assume the structure $\boxed{1}$ to be [] (empty information), and all subsumption and equality constraints of this analysis are trivially met.

In fact, a construction "$v_1$ and $v_2$" has to allow for just those complements that can follow $v_1$ by itself, but also can follow $v_2$ by itself, imposing the requirements of both verbs together. E.g., only an AP complement may follow *grew and remained*. But we also cannot assume the type requirements of the two verbs to simply get unified, because they may be inconsistent. At least, this is suggested by the following German sentences cited originally by [17] and [9].

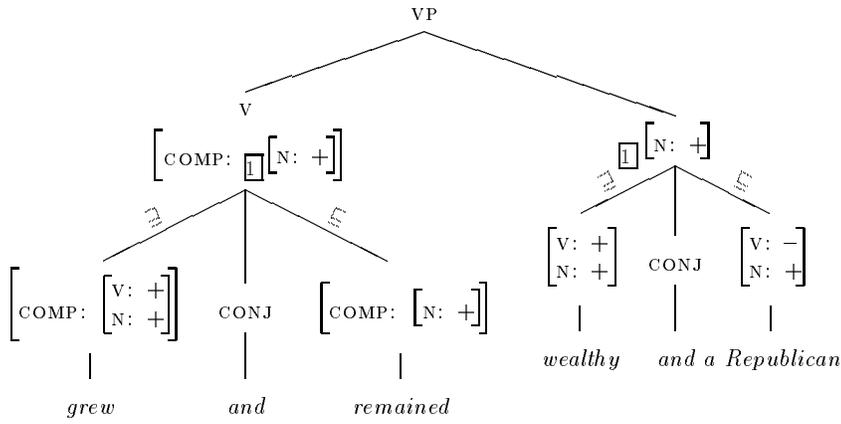

Fig. 1.  Overgeneration of the coordination rule in the unification-based setting

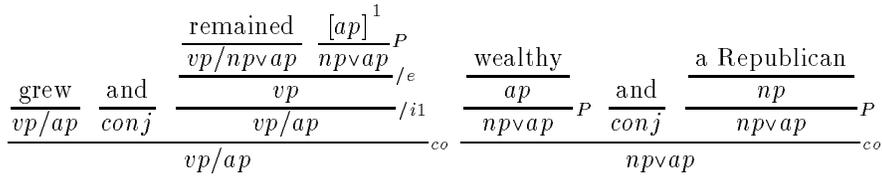

Fig. 2.  Partial proof tree showing rejection of (2) in [3]

(3)a./b./c. Er findet    und hilft      *Männer /*Kindern /  Frauen
           he find-ACC and help-DAT men-ACC  children-DAT women-ACC+DAT

Bayer and Johnson have a simple and convincing analysis accounting for this phe-
nomenon. They devise a simple extension to LCG (Lambek Categorial Grammar)
in which basic types are replaced by propositional formulas built solely from propo-
sitional variables, ∧ (conjunction) and ∨ (disjunction). Apart from the original
LCG rules only the following two rule schemata are needed:[1]

$$\frac{\phi}{\psi}P, \quad \text{if } \phi \vdash \psi \text{ in propo-} \qquad \qquad \frac{A \; conj \; A}{A} co \qquad \text{Condition: no undischarged}$$
sitional logic                                                 hypothesis in any conjunct

The first deals with propositional formulae, the second with coordination. Now, the
polymorphic verbs *become* and *remained* can simply be categorised as $vp/np\lor ap$.
A coordination like *wealthy and a Republican* receives category $np\lor ap$ by using the
weakening rule $P$ twice. However, when we want to conjoin *grew* and *remained*
using $co$, we need to strengthen the argument category of *remained* to be $ap$ (cf.
Fig. 2).   Thus the ungrammatical double coordination (2) is correctly reject-
ed.   Recasting the analysis in terms of logic, the fact that the subcategorisation
information acts as a premise of an implication — logically, functor categories
can be seen as implications — is responsible why in a coordination of verbs this

$$
\cfrac{
  \cfrac{
    \cfrac{\cfrac{\text{findet}}{vp/np\wedge acc} \quad \cfrac{[np\wedge acc\wedge dat]^{1}}{np\wedge acc}P}{vp}/e
  }{vp/np\wedge acc\wedge dat}/i1 \quad
  \cfrac{\text{und}}{conj} \quad
  \cfrac{
    \cfrac{\cfrac{\text{hilft}}{vp/np\wedge dat} \quad \cfrac{[np\wedge acc\wedge dat]^{2}}{np\wedge dat}P}{vp}/e
  }{vp/np\wedge acc\wedge dat}/i2
}{\cfrac{vp/np\wedge acc\wedge dat \qquad \cfrac{\text{Frauen}}{np\wedge acc\wedge dat}}{vp}/e}\,co
$$

Fig. 3.   The LCG analysis of (3c)

$$
\cfrac{
  \cfrac{\text{findet und hilft}}{vp/np\wedge acc\wedge dat} \quad
  \cfrac{
    \cfrac{\cfrac{\text{Männer}}{np\wedge acc}}{np\wedge (acc\vee dat)}P \quad
    \cfrac{\text{und}}{conj} \quad
    \cfrac{\cfrac{\text{Kindern}}{np\wedge dat}}{np\wedge (acc\vee dat)}P
  }{np\wedge (acc\vee dat)}co
}{}
$$

Fig. 4.   The LCG analysis showing blocking of (4)

requirement gets strengthened (the category of the coordination becomes a common weakening of all the conjoined verb categories).

Interestingly, since subcategorisation requirements are encoded as 'feature information' that must be *entailed* by the respective complement, there is no need to view different grammatical cases like *dative* and *accusative* as mutually inconsistent information as in the consistency-based approach. For instance, the German verb *helfen* may require an object to have dative case without disallowing it to be accusative as well, in other words, *helfen* would be specified as $vp/np\wedge dat$. Actually, this view of grammatical cases seems to be necessary to deal with the difference in grammaticality between (3c) and (4).

(4)  * Er findet und hilft Männer und Kindern.

*Frauen* simply needs to be of type $np\wedge acc\wedge dat$ and then the extended LCG of [3] correctly accepts (3c) (cf. Fig. 3). Also, (4) is rejected (see Fig. 4), since *Männer und Kindern* receives the weaker type $np\wedge(acc\vee dat)$, and hence does not match the combined requirement of *findet und hilft*.

## 3   Lambek calculi with subtyping

The general approach we take can be seen as adding Lambek slashes to some simple logical system, some *base logic*, for describing grammatical categories. However, seen from the outside the only effect of such base logics on the Lambek layer — so-to-speak the intercategorial layer — will be that they define orderings between 'basic' categories, more precisely subtype orderings. For example, we might want to allow for basic category descriptions like the above $np\wedge acc\wedge sg$ and $np\wedge acc$ (or if we want to employ feature logic $[np \,\&\, case{:}acc \,\&\, num{:}sg]$ and $[np \,\&\, case{:}acc]$) together with the stipulation that whenever something is of the former type, it is also of the latter. So before looking at Lambek systems over arbitrary base logics, we first give a rigorous treatment of the almost trivial extension to Lambek calculi where the set of basic types $\mathcal{B}$ is assumed to come with some subtype ordering $\preceq$. The combination schema defined later on, in which a base logic for the description of categories is employed, will be an instance of Lambek systems of this kind. Specifically, in the case of feature term descriptions, the subtype ordering will be the well-known feature term subsumption relation.

### 3.1  Syntax and semantics

Assume as given a set of basic types $\mathcal{B} = \{b_1, b_2, \ldots\}$ on which a preorder $\preceq$ (the subtype ordering) is defined, *i.e.* we require $\preceq \subseteq \mathcal{B} \times \mathcal{B}$ to be reflexive and transitive. If $b_1 \preceq b_2$, call $b_1$ a subtype of $b_2$ and $b_2$ a supertype of $b_1$.

**Definition 1** *The set of formulae $\mathcal{F}$ of $\mathsf{L}_{\preceq}$ is defined inductively by:*

*1. if $b$ is in $\mathcal{B}$, then $b$ is in $\mathcal{F}$;*

*2. if $A$ and $B$ are in $\mathcal{F}$, then $A/B$ and $B \backslash A$ are in $\mathcal{F}$.*

**Definition 2** *A (string-semantic) model for $\mathsf{L}_{\preceq}$ is a Kripke model $\langle W, \cdot, \llbracket \cdot \rrbracket \rangle$, where*

*1. $W$ is a nonempty set;*

*2. $\cdot : W \times W \mapsto W$ is an associative operation; and*

*3. $\llbracket \cdot \rrbracket : \mathcal{F} \mapsto 2^W$ maps formulae to subsets of $W$ and satisfies:*

  *a) $\llbracket b_1 \rrbracket \subseteq \llbracket b_2 \rrbracket$ whenever $b_1 \preceq b_2$;*

  *b) $\llbracket A/B \rrbracket = \{x \mid \forall y \in \llbracket B \rrbracket \ (x \cdot y \in \llbracket A \rrbracket)\}$; and*

  *c) $\llbracket B \backslash A \rrbracket = \{y \mid \forall x \in \llbracket B \rrbracket \ (x \cdot y \in \llbracket A \rrbracket)\}$.*

Let us in parallel to $\mathsf{L}_{\preceq}$ consider the logics resulting from extending with subtyping the variations $\mathsf{NL}$, $\mathsf{LP}$ and $\mathsf{NLP}$ of Lambek's calculus. Here $\mathsf{N}$ abbreviates "nonassociative" and $\mathsf{P}$ abbreviates "with permutation". We define models for $\mathsf{NL}_{\preceq}$ [respectively $\mathsf{LP}_{\preceq}$; respectively $\mathsf{NLP}_{\preceq}$] as in Definition 2, but simply allowing (requiring) the second component $\cdot$ of models to be an arbitrary [respectively commutative and associative; respectively commutative] operation.

### 3.2  Gentzen calculi

There is a well-known common formulation of the proof system of all these Lambek logic variants in form of a Gentzen sequent calculus. The extension needed to cover subtyping is the same for all variants and almost trivial.

In Gentzen calculi claims of the form $U \Rightarrow A$ are derived, which can be read as "formula $A$ is derivable from the structured database $U$", where a structured database (or G-term) is built from formulae using the single (binary) structural connective $\odot$.

The denotation mappings $[\![\cdot]\!]$ in our models are naturally extended to G-terms by defining (symbols $U$ and $V$ will henceforth always denote G-terms)

$$[\![U \odot V]\!] := \{z \in W \mid \exists x, y \in W \ (z = x \cdot y \wedge x \in [\![U]\!] \wedge y \in [\![V]\!])\}.$$

A *sequent* in each of our logics is a pair $(U, A)$, written $U \Rightarrow A$, where $U$ is a G-term and $A \in \mathcal{F}$. We call a sequent $U \Rightarrow A$ *valid*, if for all string models $\mathcal{M} = \langle W_{\mathcal{M}}, \cdot^{\mathcal{M}}, [\![\cdot]\!]^{\mathcal{M}} \rangle$, $[\![U]\!]^{\mathcal{M}} \subseteq [\![A]\!]^{\mathcal{M}}$.

$$(Ax) \ \frac{}{b_1 \Rightarrow b_2} \qquad \text{if } b_1 \preceq b_2$$

$$(/L) \ \frac{V \Rightarrow B \quad U[A] \Rightarrow C}{U[A/B \odot V] \Rightarrow C} \qquad (/R) \ \frac{U \odot B \Rightarrow A}{U \Rightarrow A/B}$$

$$(\backslash L) \ \frac{V \Rightarrow B \quad U[A] \Rightarrow C}{U[V \odot B \backslash A] \Rightarrow C} \qquad (\backslash R) \ \frac{B \odot U \Rightarrow A}{U \Rightarrow B \backslash A}$$

$$(Cut) \ \frac{V \Rightarrow A \quad U[A] \Rightarrow C}{U[V] \Rightarrow C}$$

Fig. 5.   Gentzen calculus $\mathsf{NL}_{\preceq}$

Now the logical rules for all variant systems are the same and shown in Fig. 5. The notation $U[X]$ in rules $/L$, $\backslash L$ and $Cut$ stands for a G-term $U$ with a distinguished occurrence of the (sub-)G-term $X$, and by using $U[Y]$ in the same rule we denote the result of substituting $Y$ for that occurrence $X$ in $U[X]$. The variations between the four logics comes from their different use of the so-called structural rules ASSOCIATIVITY and PERMUTATION, by which the G-term connective $\odot$ may be forced to obey the appropriate combination of being associative or not and being commutative or not. However, we do not assume these rules explicitly, but rather, as it is standard, let them be implicit in the notion of G-term for the respective logic by taking G-terms modulo the respective combination of these rules. For instance, in $\mathsf{L}$ connective $\odot$ is assumed to be associative, which allows us to see G-terms simply as (nonempty) sequences (strings) of formulas with $\odot$ denoting concatenation.

The only departure from standard Lambek calculus is the axiom scheme. It is the straightforward generalisation needed to generate all valid sequents of two basic formulae.

### 3.3 SOUNDNESS AND COMPLETENESS

Let us now check that our proof systems are indeed sound and complete. In the following completeness theorem the case of $\mathsf{L}_{\preceq}$ already follows from an argument given in [7], showing that the product-free Lambek calculus (*i.e.* $\mathsf{L}_{\preceq}$ where $\preceq$ is the identity) augmented by a set of additional axiom schemata $R$ is sound and complete w.r.t. the class of string-semantic models satisfying $R$. The proof here is a straightforward adaptation of this argument, however we give it for all Lambek variants simultaneously.

**Theorem 3** $U \Rightarrow A$ *is derivable in* $\mathsf{L}_{\preceq}$ *[resp.* $\mathsf{NL}_{\preceq}$, $\mathsf{LP}_{\preceq}$, $\mathsf{NLP}_{\preceq}$*] iff* $U \Rightarrow A$ *is valid (with the respective condition on the structural connective* $\odot$*).*

PROOF. The direction from left to right (soundness) is shown by the usual induction on length of proofs. Validity of axiom instances directly follows from model condition (3a). For the other direction, we construct the canonical model $\mathcal{CM}$ as follows. Let $\mathcal{CM} = \langle W, \odot, \llbracket \cdot \rrbracket \rangle$ with $\langle W, \odot \rangle$ being the free algebra generated by $\odot$ over the formulas (modulo the respective structural rules) and let $\llbracket \cdot \rrbracket$ be defined as

$$\llbracket A \rrbracket := \{ u \in W \mid \; \vdash u \Rightarrow A \}.$$

Here $\vdash \Gamma$ stands for $\Gamma$ *is derivable in the respective logical system*. Observe that due to the reflexivity of $\preceq$ all sequents derivable in the original Lambek calculus (resp. in its variants $\mathsf{NL}$, $\mathsf{LP}$, $\mathsf{NLP}$) are also derivable here.

1. $\mathcal{CM}$ is a model: we check the conditions *(a)–(c)* in the definition of $\llbracket \cdot \rrbracket$.

    a) Suppose $b_1 \preceq b_2$. Then $U \in \llbracket b_1 \rrbracket$ iff $\vdash U \Rightarrow b_1$, but then with *Cut* on $b_1 \Rightarrow b_2$, $\vdash U \Rightarrow b_2$.
    b) We need to show for all $U \in W$, $A, B \in \mathcal{F}$:

    $$\vdash U \Rightarrow A/B \;\; \text{iff} \;\; \forall V \in W \; (\vdash V \Rightarrow B \; \text{implies} \; \vdash U \odot V \Rightarrow A).$$

    So, suppose for arbitrary $U, A, B$, and $V$, $\vdash U \Rightarrow A/B$ and $\vdash V \Rightarrow B$. The latter gives, using $/L$ on $A \Rightarrow A$, $A/B \odot V \Rightarrow A$, and hence with *Cut*, $\vdash U \odot V \Rightarrow A$. For the converse assume $U, A, B$ such that the condition of the right-hand side holds. With $V = B$ we then get $\vdash U \odot B \Rightarrow A$, since $\vdash B \Rightarrow B$ is true. But then, by $/R$ we derive the required $U \Rightarrow A/B$.
    c) symmetrical

2. $\mathcal{CM}$ invalidates underivable sequents: for suppose $\nvdash U \Rightarrow A$, hence $U \notin \llbracket A \rrbracket$. But clearly $U \in \llbracket U \rrbracket$ contradicting the validity of $U \Rightarrow A$.

This completes the proof.                                                                          □

## 3.4   SOME PROPERTIES OF THE CALCULI

It is helpful to classify the occurrences of subformulas in $\mathcal{F}$-formulas and sequents into positive (polarity $= +1$) and negative (polarity $= -1$) as follows.

## Definition 4 (positive/negative subformula occurrences)

– *A occurs* positively *in A.*

– *If the polarity of B in A is p and B has the form $C/D$ or $D\backslash C$, then the polarity of that occurrence of C [resp. D] in A is p [resp. −p].*

Next we extend the subtyping relation $\preceq$ to complex types by stipulating

$$\text{if } A \preceq A' \text{ and } B' \preceq B \text{ then } A/B \preceq A'/B' \text{ and } B \backslash A \preceq B' \backslash A'. \tag{1}$$

Evidently, with this definition we remain faithful to the intended meaning of subtyping, namely that it denotes the subset relation.

**Lemma 5** *If $A \preceq B$, then for any $\mathsf{L}_{\preceq}$ model $\llbracket A \rrbracket \subseteq \llbracket B \rrbracket$.*

Using completeness we trivially obtain

**Theorem 6** *If $A \preceq B$, then $\vdash A \Rightarrow B$.*

The following two derived rules make explicit the monotonicity properties of derivability with respect to subtyping on the left- and on the right-hand side. They are specialisations of *Cut* taking into account Theorem 6. Be warned that the notation $U[A]$ stands for $U$ with a *G-term occurrence $A$* (not an arbitrary subformula).

$$(strengthen\ L)\ \frac{U[B] \Rightarrow C}{U[A] \Rightarrow C} \quad \text{if } A \preceq B \qquad (weaken\ R)\ \frac{U \Rightarrow A}{U \Rightarrow B} \quad \text{if } A \preceq B$$

### 3.4.1 Cut elimination

An important result is that the *Cut* rule is redundant, since from this decidability (in fact even $\mathcal{NP}$-easyness) directly follows. *Cut*-free proofs have the nice property that the length of the proof is bounded linearly by the size of the sequent to be proven. So, let us in the following call $\mathsf{NL}_{\preceq}^{-}$ the system comprising of the rules and axiom shown in Fig. 5, except for the *Cut*-rule.

**Theorem 7 (Cut Elimination)** *$U \Rightarrow A$ is derivable in $\mathsf{NL}_{\preceq}$ iff it is derivable in $\mathsf{NL}_{\preceq}^{-}$.*

(A proof is given in the appendix.)

### 3.4.2 Context-freeness

We show here that the addition of subtyping to the calculus of Lambek categorial grammars does not extend their generative capacity. A categorial grammar based on $\mathsf{L}_{\preceq}$ can always be compiled out into a (possibly much larger) Lambek categorial grammar (and hence, also into a context-free grammar).

Assume as given a system $\mathsf{L}_{\preceq} = (\mathcal{F}, \vdash_{\mathsf{L}_{\preceq}})$ and a finite alphabet $\mathcal{T}$ of lexical entities.

**Definition 8** *An $\mathsf{L}_{\preceq}$ grammar $G$ for $\mathcal{T}$ is a pair $(\alpha, S)$ consisting of the lexical assignment $\alpha : \mathcal{T} \mapsto 2^{\mathcal{F}}$, such that for any $t \in \mathcal{T}$, $\alpha(t)$ is a finite set of types, and the distinguished (sentence) type $S \in \mathcal{F}$.*
*The language generated by $G$ is*

$$L(G) := \{t_1 \dots t_n \in \mathcal{T}^* \mid \exists B_1 \dots B_n\ (\vdash_{\mathsf{L}_{\preceq}} B_1 \dots B_n \Rightarrow S \wedge \bigwedge_{i=1}^{n} B_i \in \alpha(t_i))\}$$

**Theorem 9 (Context-freeness)** *For any $\mathsf{L}_{\preceq}$ grammar $G$ the language $L(G)$ is context-free.*

(A proof is given in the appendix.)

### 4　Layering Lambek over some base logic

We now describe how the abstract placeholder of the subtype ordering is sensibly filled in in constraint-based Lambek grammar. As explained above we assume some specification language (or base logic) whose purpose it is to allow for more fine-grained descriptions of the basic categories. Before we proceed to describe a concrete choice of base logic, let us first consider abstractly what we demand of such a logic and how an integration with Lambek logic can be defined generally in a way exhibiting the desired properties.

Formally, we assume a base logic $\mathsf{BL}$ consisting of:

1. the denumerable set of formulae $\mathcal{BF}$

2. the class $\mathcal{C}$ of models $\flat$ of the form $\langle W_\flat, [\![\cdot]\!]^\flat \rangle$, where:

   a) $W_\flat$ is some nonempty set
   b) $[\![\cdot]\!]^\flat : \mathcal{BF} \mapsto 2^{W_\flat}$ maps formulae to subsets of $W_\flat$

We will be interested in logical consequence between two single formulae, $i.e.$, in the relation $\phi \models_{\mathsf{BL}} \psi \Leftrightarrow_{\mathsf{def}} \forall \flat \in \mathcal{C} \, ([\![\phi]\!]^\flat \subseteq [\![\psi]\!]^\flat)$ for all $\phi, \psi \in \mathcal{BF}$. We also assume a sound and complete deduction system $\vdash_{\mathsf{BL}}$ for answering this question. Note that $\models_{\mathsf{BL}}$ is a preorder. Now we define the layered Lambek logic $\mathsf{L}(\mathsf{BL})$ over some given base logic $\mathsf{BL}$ as follows.

**Definition 10** *The logic* $\mathsf{L}(\mathsf{BL})$ *is the logic* $\mathsf{L}_{\models_{\mathsf{BL}}}$ *over basic types* $\mathcal{BF}$.

This means, we let $\mathsf{L}(\mathsf{BL})$ simply be a subtyping Lambek logic $\mathsf{L}_{\preceq}$ taking as $\preceq$ the consequence relation over $\mathsf{BL}$-formulae. Thus we inherit the complete and decidable (provided $\models_{\mathsf{BL}}$ is decidable) proof system of $\mathsf{L}_{\preceq}$, but also its limited generative capacity.

Note that every formula of $\mathsf{L}(\mathsf{BL})$ has an outer (possibly empty) Lambek part, which composes some $\mathsf{BL}$-formulae with the connectives $/$ and $\backslash$, but no $\mathsf{BL}$ connective may scope over formulae containing $/$ or $\backslash$ (hence the term 'layered logic').

The reader will have noticed that Bayer and Johnson's logic presented in Section 2 is a layered logic of this kind, where $\mathsf{BL}$ is instantiated to be propositional logic (restricted to connectives $\wedge$ and $\vee$).

It may be a bit surprising that the generative capacity is not affected by the choice of the base logic. Even if we admit full standard first-order logic, we do not exceed context-free power. We see a cause of this in the fact that there is no direct interrelation between $\mathsf{L}(\mathsf{BL})$-models and base models. Only on the abstract level of the consequence relation $\models_{\mathsf{BL}}$ the notion of base model enters the conditions on $\mathsf{L}(\mathsf{BL})$-models.[2] On the other hand $\mathsf{L}(\mathsf{BL})$ enjoys a particularly simple and modular

---

[2]　As an alternative to our model conception, consider the option of defining an $\mathsf{L}(\mathsf{BL})$-model in a such way that it contains a base model as a part. For instance, when given a base model $\flat$, we could stipulate

$$[\![\phi]\!] = [\![\phi]\!]^\flat \text{ for all } \phi \in \mathcal{BF}.$$

proof system. Proofs can be built by decomposing the Lambek part of a goal sequent with the usual rules until one reaches pure $\mathsf{BL}$-sequents (Lambek axioms). Those are then proved in pure $\vdash_{\mathsf{BL}}$ by *independent* proofs for each sequent, *i.e.*, the proof system $\vdash_{\mathsf{BL}}$ can be considered a black box.

## 4.1 Feature logic as base logic

We now consider the consequence of choosing feature logic as the base logic. For this variant of the logic to be complete we need a proof system for deciding $\phi \models_{\mathsf{FL}} \psi$ *i.e.* checking whether $\phi$ *entails* $\psi$. Proof systems for deciding *entailment* between feature constraints are well known (cf. [2] [1]).

We provide a simple entailment checking for a restricted feature logic with the following syntax:

$$\phi, \psi \longrightarrow \quad x \qquad \text{variable}$$
$$a \qquad \text{atom}$$
$$\top \qquad \text{top}$$
$$\bot \qquad \text{bottom}$$
$$f : \phi \qquad \text{feature term}$$
$$\exists x(\phi) \qquad \text{existentially quantified variable}$$
$$\phi \ \& \ \psi \qquad \text{conjunction}$$

Since the semantics of feature logic is well known, we do not provide a model theoretic semantics here (cf. [19] for details).

As a first step in determining the entailment $\phi \models_{\mathsf{FL}} \psi$ we translate the constraints $x = \phi$ and $x = \psi$ into normal form by employing normalisation procedure for feature logic (see [19]). A normal form translation of the constraint $x = \phi$ results in a constraint of the form $\exists x_0 \ldots x_n \ \phi'$ where $\phi'$ is a conjunction of simple constraints. In our case, these are of the form : $x = a, x = \top, x = \bot, x = f : y$. Thus, normal form translation of $x = (cat : np \ \& \ case : acc)$ gives $\exists y_0 y_1 (x = cat : y_0 \ \& \ y_0 = np \ \& \ x = case : y_1 \ \& \ y_1 = acc)$. The normal form translation, apart from providing a simplified conjunction of constraints, also decides consistency of the initial formula.

Now we describe a simple method for deciding $\phi \models_{\mathsf{FL}} \psi$ in the general case where both $\phi$ and $\psi$ are consistent. Let $\exists x_0 \ldots x_n \ \phi'$ (resp. $\exists y_0 \ldots y_m \ \psi'$) denote the normal form translation of $x = \phi$ (resp. $x = \psi$). For simplicity of presentation, we assume that the existentially quantified variables $x_0 \ldots x_n$ and $y_0 \ldots y_m$ are disjoint. Since the existential quantification in $\exists x_0 \ldots x_n \ \phi'$ is not needed in the simplification procedure, we remove this from consideration. We then apply the entailment checking procedure given in Figure 6 to $\phi' \models_{\mathsf{FL}} \exists y_0 \ldots y_m \ \psi'$ calling $\phi'$ the *context* and $\exists y_0 \ldots y_m \ \psi'$ the *guard*.

---

instead of condition *(a)* on $[\![ \cdot ]\!]$ (assume $W_s \subseteq W$). (In our model conception such models are allowed, but not required). A consequence of this option is that certain logical behaviour of $\mathsf{BL}$-formulae may get exported to the Lambek level leading to incompleteness there. E.g., when $\mathsf{BL}$ contains a constant (or formula), say FALSE, denoting the empty set, any sequent of the form $U[\text{FALSE}] \Rightarrow A$ will be valid.

$$(SAtom) \quad \frac{x = a \ \& \ \phi \models_{\mathsf{FL}} \exists y_0 \ldots y_n \ \psi}{x = a \ \& \ \phi \models_{\mathsf{FL}} \exists y_0 \ldots y_n \ x = a \ \& \ \psi}$$

$$(SFeat) \quad \frac{x = f : y \ \& \ \phi \models_{\mathsf{FL}} \exists y_0 \ldots y_n \ \psi}{x = f : y \ \& \ \phi \models_{\mathsf{FL}} \exists y_0 \ldots y_n \ x = f : y \ \& \ \psi}$$

$$(SFeatExist) \quad \frac{x = f : y \ \& \ \phi \models_{\mathsf{FL}} \exists y_0 \ldots y_i \ldots y_n \ [y/y_i]\psi}{x = f : y \ \& \ \phi \models_{\mathsf{FL}} \exists y_0 \ldots y_i \ldots y_n \ x = f : y_i \ \& \ \psi}$$

Fig. 6.   Entailment checking in simple feature logic

The rules given in Figure 6 are to be read from bottom to top. These rules simplify the guard with respect to the context. The rules $SAtom$ and $SFeat$ are self-explanatory. In rule $SFeatExist$ the notation $[y/y_i]\psi$ means replacing every occurrence of $y_i$ with $y$ in $\psi$. Once the entailment checking rules terminate, entailment and disentailment can be decided by inspection.

- $\phi \models_{\mathsf{FL}} \psi$ if entailment checking of $\phi' \models_{\mathsf{FL}} \exists y_0 \ldots y_m \ \psi'$ simplifies $\psi'$ to a possibly empty conjunction of formula of the form $y_i = \top$.

- $\phi \not\models_{\mathsf{FL}} \psi$ (*i.e.* $\phi$ *disentails* $\psi$) if entailment checking of $\phi' \models_{\mathsf{FL}} \exists y_0 \ldots y_m \ \psi'$ simplifies to the form $x = a \ \& \ \phi' \models_{\mathsf{FL}} \exists y_0 \ldots y_m \ x = \tau \ \& \ \psi'$ [or the form $x = \tau \ \& \ \phi' \models_{\mathsf{FL}} \exists y_0 \ldots y_m \ x = a \ \& \ \psi'$] where $\tau$ is one of:

    · $b$ with $a, b$ distinct
    · $f : z$

- $\phi \models_{\mathsf{FL}} \psi$ is *blocked* if neither of the above two conditions hold.

This completes the basic building blocks needed to implement a proof system for a Lambek calculus with feature logic as the base logic.

## 5   Discussion

The initial motivation for conducting this research was the lack of a model-theoretic semantics for unification-based versions of categorial grammar, standing in contrast to the apparently clear intuition of what categorial types over feature terms should mean. We were guided by the insight (or basic assumption) that a functor type $A/B$, according to the traditional semantics, may always be applied to subtypes of $B$ (yielding an $A$), but not necessarily to supertypes. Combined with the idea that feature terms as basic types essentially provide a means to express fine-structuring of types, *i.e.*, subtyping, this led us to devise a simple model theory embodying just those assumptions, which however is accompanied with a subsumption-based (though equally simple) proof system. This work in a sense complements work by Dörre, König and Gabbay [4], in which a model-theoretic counterpart of the unification-based proof system is constructed using the paradigm of fibred semantics (cf. [8]).

One important issue for grammar logics upon which we have remained silent up to now is semantics construction. There was no need to address it so far, because our logic is completely neutral in that respect and does not involve any commitment to one of the two familiar approaches to construct semantics. We either can please the categorial grammar purist and use the Curry-Howard(-van Benthem) correspondence to view rules as recipes to cook up lambda terms (cf. [22]). Or we may contend aficionados of the HPSG way and employ an additional layer in which semantic formulae are built up by unification, as we will see below.

In a Lambek-van Benthem system, as is well-known, each type is paired with a semantic formula and $L$-rules trigger function application and $R$-rules lambda abstraction on these formulae. Since this does in no way interfere with our extension, we can apply this method unchanged.

On the other hand, if we were to adopt a HPSG style semantics, then variable sharing across categories is needed, *i.e.*, we cannot come by simply adding to the feature structures of basic types a SEM feature, since no information can be 'percolated' out of local trees. What we propose here is to use the combination scheme of [4] and combine $\mathsf{L(FL)}$ with another layer of *feature constraints*. In that second feature-logical layer, however, unification is employed to match categories. We describe here in short how this construction works.

*Double Layering*   To each basic type in an $\mathsf{L(FL)}$ formula associate a new variable. On these variables we then can impose (feature) constraints noted in a top-level conjunction as a third component of a sequent. E.g., an HPSG-like lexical assignment for the control verb *persuade* would be of type (unification-layer variables written as superscripts)

$$([cat\!:\!np]^X \setminus [cat\!:\!s]^S) \, / \, [cat\!:\!np]^Y \, / \, ([cat\!:\!np]^Y \setminus [cat\!:\!s \,\&\, vform\!:\!inf]^Z),$$

together with constraints on the variables $X, S, Y, Z$. For instance, we might require that $S$ is constrained to be a *persuading relation* as given by:

$$S = \left[ \text{CONTENT} \left[ \begin{array}{ll} \text{RELATION} & persuade \\ \text{INFLUENCE} & X \\ \text{INFLUENCED} & Y \\ \text{SOA--ARG} & Z \end{array} \right] \right]$$

A proof proceeds as in the system $L(FL)$, but additionally maintains as a global environment a feature constraint $\Phi$, initially the conjoined feature constraints of all formulae in the goal sequent. Whenever we apply the axiom schema on some $[b_1]^X \Rightarrow [b_2]^Y$, we add $X = Y$ to $\Phi$, normalize, and continue if the result is consistent. This simply means, we unify the feature graphs (encoded as the constraints) of $X$ and $Y$. Thus the composition of the content structure proceeds in exactly the same way as in HPSG (or other comparable unification-based frameworks).

The important point here to note is that unification-based and subsumption-based argument binding can be combined into a single logic[3] and complement each other. Speaking in HPSG terms, we have the flexibility to choose which parts of a sign we want only to be matched and which parts to be unified, when it is combined with others in a local tree.[4]

On the processing side we believe that the separation of a subsumption-based layer (of context-free power) and a unification-based layer offer similar benefits like the distinction between c-structure and f-structure in LFG. For instance, we can easily precompile the formulas of the subsumption layer into a type hierarchy of atomic symbols and thus may be able to employ efficient indexing techniques during parsing.

A final point we want to make concerns other extensions to the original Lambek calculus which appear to be prerequisites for many linguistically interesting analyses. By that we mean for example Moortgat's non-directional slash operator $\uparrow$, allowing for non-peripheral gaps, his operator $\Uparrow$ for generalised quantifiers, Morrill's multimodal and discontinuity operators, structural modalities etc. (cf. [13]). We believe that our extension of subtyping is well compatible with (at least most of) these additional devices and offers an orthogonal extension to these.

## 6    Conclusion

We have shown that a simple and happy marriage between constraint-based grammars and categorial grammars is technically feasible with an appealingly simple model theory. Our hybrid grammar logic permits extant categorial proof systems to be carried over in the new system. Furthermore, the logic is parameterised over arbitrary (constraint) logics as long as a reasoning mechanism for determining

---

[3] in much the same way as we can add a unification component for feature structures to a context-free grammar to obtain LFG.

[4] The reader might wonder whether there is a simpler way to allow for information percolation through variable binding. Suppose we would use the strategy "after having checked that the category serving as actual argument (the complement sign) is subsumed by the functor type's argument description (the respective slot of the subcat list of the head), just unify the two". Hence, variables in the subsuming type would be bound, possibly carrying that information to other parts of the functor type's structure. For instance, we could have a modifier type $\exists x[cat\!:\!x/cat\!:\!x]$ or a coordination type $\exists x[cat\!:\!x\backslash cat\!:\!x/cat\!:\!x]$, the result type of which would depend on the type(s) of its argument(s). But consider what happens, when we apply that coordination type to two arguments of different types:

$$cat\!:\!(npvap) \odot \exists x[cat\!:\!x\backslash cat\!:\!x/cat\!:\!x] \odot cat\!:\!np$$

Using $/L$ and $\backslash L$ this reduces to $cat\!:\!x$ plus the two sequents $cat\!:\!np \Rightarrow cat\!:\!x$ and $cat\!:\!(npvap) \Rightarrow cat\!:\!x$. Now, if we choose to first prove the first, $x$ gets bound (globally) to $np$ and the second sequent fails, but choosing the other order $x = (npvap)$ and the proof succeeds. This means that our naïve proof procedure is sensitive to the order of rule application. In other words, to guarantee the completeness we would have to *search* for a particular sequence of rule ordering. This is an undesirable situation that we want to avoid, since it will result in a vastly inefficient proof procedure.

In addition there is a *semantic problem* namely that it becomes rather difficult to provide a sensible semantics to categories $B\backslash A$ and $A/B$. In particular, what we witnessed in the example above is that the denotation of $A$ in $B\backslash A$ (or $A/B$) is going to be dynamic and contingent on what $A$ actually unifies with.

entailment and consistency is provided. We believe that crucial to the success of this approach is the novel use of subsumption (or entailment) checking as opposed to just unification.

## Appendix

### Proofs of Theorems of Section 3.4

CUT ELIMINATION

The following lemma considers derivability of a special case of each of the two derived rules above in $\mathsf{NL}^-_{\preceq}$, laying the seed for the *Cut* elimination proof. It is stated as well as the *Cut* elimination theorem with respect to derivability in $\mathsf{NL}_{\preceq}$, the weakest of the four systems, but it should be kept in mind that these facts about derivability hold a fortiori in the other systems as well.

**Lemma 11** *If $U[b_2] \Rightarrow C$ [respectively $U \Rightarrow b_1$] is derivable in $\mathsf{NL}^-_{\preceq}$ and $b_1 \preceq b_2$, then $U[b_1] \Rightarrow C$ [respectively $U \Rightarrow b_2$] is derivable in $\mathsf{NL}_{\preceq}$.*

PROOF. By induction on the length of the proof of $\Gamma = U[b_2] \Rightarrow C$.

$n = 0$: Then $U = b_2$ and $C = b_3 \in \mathcal{BF}$ [resp. $U = b_3$]. By transitivity of $\vdash_{\mathsf{BL}}$, $b_1 \Rightarrow b_3 \;(= U[b_1] \Rightarrow C)$ [resp. $U \Rightarrow b_1$] is an axiom.

$n > 0$: We distinguish cases according to the last rule used in the proof of $\Gamma$ and state the bracketed cases separately:

/$R$: Then $C = A/B$ and there is a proof of $(U[b_2], B) \Rightarrow A$ of smaller length. Hence by induction hypothesis (IH) we get $(U[b_1], B) \Rightarrow A$ and by /$R$ the required sequent.

[/$R$ :] cannot be last step

/$L$: $U$ contains distinguished occurrences of the two G-terms $b_2$ and $(A/B, V)$. Hence, either $b_2$ occurs in $V$, in which case by IH on $V \Rightarrow B$ the claim is shown, or $U = U[(A/B, V), b_2]$, *i.e.* $b_2$ occurs also in $U[A]$ as a G-term different from $A$. In this case the claim follows via IH on $U[A, b_2] \Rightarrow C$.

[/$L$ :] obvious, since we get by IH $U[A] \Rightarrow b_2$.

$\backslash R$, $\backslash L$, [$\backslash R$], and [$\backslash L$] are completely analogous. □

**Proof of the Cut Elimination Theorem.** We point out what needs to be changed in the Cut elimination proof for standard Lambek calculus (cf., e.g., [5]). Let the *degree* of an application of *Cut* be the number of occurrences of connectives in the cut-formula $A$, the formula eliminated by applying *Cut*. Then the standard argument goes by showing that whenever a sequent $\Gamma$ has a proof which contains exactly one application of *Cut*, which is of degree $d$, and this is the last step in that proof, *i.e.* the proof has the form

$$(Cut) \quad \frac{\overset{\vdots}{\Gamma_1} \quad \overset{\vdots}{\Gamma_2}}{\Gamma}$$

then by a case analysis of the two steps introducing $\Gamma_1$ and $\Gamma_2$, it follows that $\Gamma$ is derivable from the premises of those steps involving either no $Cut$ or only $Cut$ applications of lower degree. The arguments for all the cases carry over identically to our system except for the case where one of $\Gamma$'s premises is an axiom. So, suppose $\Gamma_1 = b_1 \Rightarrow b_2$ with $b_2$ being the $Cut$-formula. Then $\Gamma_2 = U[b_2] \Rightarrow C$, and whence by Lemma 11, $\Gamma = U[b_1] \Rightarrow C$ has a cut-free proof. But also if $\Gamma_2 = b_1 \Rightarrow b_2$ with $b_1$ being cut, we get $\Gamma_1 = V \Rightarrow b_1$ and again by Lemma 11 the claim holds.                                                                                      □

### Context-freeness

The following charaterisation of $\preceq$, which is a simple consequence of its definition for complex types, will be useful in proving the context-freeness theorem.

**Lemma 12** $A \preceq B$ *iff* $B$ *is the result of substituting in* $A$ *0 or more negative subformulae occurrences* $b_1, \ldots, b_n$ *of basic types by subtypes and 0 or more positive subformulae occurrences* $b_1', \ldots, b_m'$ *by supertypes.*

**Proof of the Context-freeness Theorem.** We show how to construct for an arbitrary $\mathsf{L}_{\preceq}$ grammar $G = (\alpha, S)$ a finite set of (pure) Lambek grammars such that $L(G)$ is the union of the languages generated by these. Since Lambek grammars generate only context-free languages [14] and context-free languages are closed under union, $L(G)$ is context-free.

Let $\mathcal{B}|_{\alpha}$ be the (finite) subset of $\mathcal{B}$ of basic formulae occurring (as subformulae) in some $\alpha(t_i)$. Call for arbitrary $A \in \mathcal{F}$ $super^+(A)$ (resp. $super^-(A)$) the set of formulae $A'$ such that $A'$ is the result of replacing in $A$ 0 or more positive (resp. negative) subformula occurrences $b_1, \ldots, b_n$ of basic types by respective supertypes from $\mathcal{B}|_{\alpha}$. For instance, if $\mathcal{B}|_{\alpha} = \{b_1, b_2\}$ where $b_1 \prec b_2$ then $super^+(b_1/(b_1/b_1)) = \{b_1/(b_1/b_1), b_1/(b_1/b_2), b_2/(b_1/b_1), b_2/(b_1/b_2)\}$. We now define

$$\overline{S} := super^-(S) = \{S_1, \ldots, S_m\}$$
$$\overline{\alpha}(t) := \{B' \mid \exists B \ B \in \alpha(t), B' \in super^+(B)\}$$

The Lambek grammars sought for are $G_i = (\overline{\alpha}, S_i)$ (over the Lambek type system $\mathsf{L}_{Id}$) for $i = 1, \ldots, m$. We let $\overline{L}$ stand for the union of their languages and show:

$\overline{L} \subseteq L(G)$: Suppose $w = t_1, \ldots, t_n \in \overline{L}$. Then there exists a $j$ such that $w$ is generated by $G_j$, and hence there are $B_1', \ldots, B_n'$ such that $B_i' \in \overline{\alpha}(t_i)$ and $\vdash_{\mathsf{L}_{Id}} B_1' \ldots B_n' \Rightarrow S_j$. But then there are $B_1, \ldots, B_n$ such that $B_i' \in super^+(B_i)$ and $B_i \in \alpha(t_i)$. Since valid derivations in $\mathsf{L}_{Id}$ remain valid in $\mathsf{L}_{\preceq}$ we get $\vdash_{\mathsf{L}_{\preceq}} B_1' \ldots B_n' \Rightarrow S_j$ and then with $(strengthen\ L)$ and $(weaken\ R)$ (due to $B_i \preceq B_i'$ and $S_j \preceq S$, cf. Lemma 12) also $\vdash_{\mathsf{L}_{\preceq}} B_1 \ldots B_n \Rightarrow S$, i.e., $w \in L(G)$.

$L(G) \subseteq \overline{L}$: Suppose $w \in L(G)$, *i.e.*, there are $B_1, \dots, B_n$ such that $\vdash_{\mathsf{L}_{\preceq}}$ $B_1 \dots B_n \Rightarrow S$ and $B_i \in \alpha(t_i)$. Assume $\Delta$ is a proof of this sequent in $\mathsf{L}_{\preceq}$ and relies on the axiom instances $b_1 \Rightarrow b_1', \dots, b_k \Rightarrow b_k'$ (*i.e.* $b_i \preceq b_i'$). If we replace these instances by $b_1' \Rightarrow b_1', \dots, b_k' \Rightarrow b_k'$, but keep the rest of the proof structure, we obtain an $\mathsf{L}_{Id}$ proof of a sequent $B_1' \dots B_n' \Rightarrow S'$, where $B_i' \in super^+(B_i)$ for $i = 1, \dots, n$ and $S' \in super^-(S)$ (note that the left-hand sides of axioms appear as subformulae in positive occurrences on left-hand sides or negative occurrences on right-hand sides of derived sequents). Hence $B_i' \in \overline{\alpha}(t_i)$ for all $1 \leq i \leq n$ and $S' = S_j$ for some $1 \leq j \leq m$, implying that $w$ is generated by $G_j$.                                   □